\documentclass[11pt]{article}

\usepackage{hyperref}
\usepackage{multirow}
\usepackage{amssymb,amsmath,amsthm}
\usepackage{latexsym,amsfonts}
\usepackage{color}
\usepackage{graphicx}
  \DeclareGraphicsExtensions{.jpeg,.png,.jpg,.eps,.pdf}
\usepackage{tikz}
\usepackage{epstopdf}

\def\a{{\bf a}}

\def\f{{\bf f}}

\def\h{{\bf h}}

\def\rbold{{\bf r}}

\def\v{{\bf v}}

\def\y{{\bf y}}

\def\B{{\cal B}}
\def\C{{\cal C}}

\def\F{{\cal F}}

\def\K{{\cal K}}
\def\KR{{\cal KR}}

\def\R{{\mathbb R}}

\def\al{\alpha}
\def\d{\delta}

\def\e{\epsilon}
\def\g{\gamma}

\def\l{\lambda}

\def\om{\omega}
\def\OM{\Omega}
\def\r{\rho}
\def\s{\sigma}

\def\t{\tau}
\def\th{\theta}

\def\boldeta{{\boldsymbol \eta}}

\def\bth{{\boldsymbol \theta}}
\def\bphi{{\boldsymbol \phi}}

\def\bxi{{\boldsymbol \xi}}

\def\tai{t \ap \infty}

\def\ap{\rightarrow}

\def\seq{\subseteq}

\def\bz{{\bf 0}}

\def\fa{\; \forall}

\def\half{\frac{1}{2}}

\def\as{\mbox{ a.s.}}

\def\nm{\Vert}

\renewcommand{\and}{\mbox{$\wedge$}}

\newcommand{\bc}{\begin{center}}
\newcommand{\ec}{\end{center}}
\newcommand{\be}{\begin{equation}}
\newcommand{\ee}{\end{equation}}
\newcommand{\bd}{\begin{displaymath}}
\newcommand{\ed}{\end{displaymath}}
\newcommand{\ba}{\begin{array}}
\newcommand{\ea}{\end{array}}
\newcommand{\ben}{\begin{enumerate}}
\newcommand{\een}{\end{enumerate}}
\newcommand{\bit}{\begin{itemize}}
\newcommand{\eit}{\end{itemize}}
\newcommand{\beq}{\begin{eqnarray}}
\newcommand{\eeq}{\end{eqnarray}}
\newcommand{\btab}{\begin{tabular}}
\newcommand{\etab}{\end{tabular}}
\newcommand{\bfig}{\begin{figure}}
\newcommand{\efig}{\end{figure}}
\newcommand{\btp}{\begin{tikzpicture}}
\newcommand{\etp}{\end{tikzpicture}}

\def\gV{\nabla V}

\newcommand{\nmm}[1]{ \nm #1 \nm }
\newcommand{\nmeu}[1]{ \nm #1 \nm_2 }
\newcommand{\nmeusq}[1]{ \nm #1 \nm_2^2 }
\newcommand{\nmi}[1]{ \nm #1 \nm_\infty}

\newcommand{\nmS}[1]{ \nm #1 \nm_S }

\newcommand{\dx}[2]{\frac{d#1}{d#2}}
\newcommand{\px}[2]{\frac{\partial #1}{\partial #2}}
\newcommand{\tpx}[3]{\frac{\partial^2 #1}{\partial #2 \partial #3}}

\newcommand{\IP}[2]{ \langle #1 , #2 \rangle }

\newcommand{\halmos}{\hfill $\Box$}

\def\nmsl1{\nm_{{\rm SL1}}}

\newcommand{\bths}{\bth^*}
\def\sb{\mathbf{s}}
\def\Vd{\dot{V}}

\definecolor{verm}{rgb}{0.6,0.2,0.2}
\definecolor{purp}{rgb}{0.3,0.1,0.6}
\definecolor{purple}{rgb}{0.4,0.0,0.6}
\definecolor{bggreen}{rgb}{0.1,0.3,0.1}
\definecolor{dgreen}{rgb}{0.1,0.6,0.1}
\definecolor{black}{rgb}{0.0,0.0,0.0}
\definecolor{crim}{rgb}{0.3,0.1,0.1}
\definecolor{dred}{rgb}{0.5,0.1,0.1}

{\bf}{\it}
\newtheorem{definition}{Definition}{\bf}{\it}
\newtheorem{example}{Example}{\bf}{\rm}
{\bf}{\it}
\newtheorem{theorem}{Theorem}{\bf}{\it}
{\bf}{\it}
{\bf}{\it}
{\bf}{\rm}

\begin{document}

\title{
Convergence of Stochastic Approximation via\\
Martingale and Converse Lyapunov Methods
}

\author{M.\ Vidyasagar
\thanks{
The author is SERB National Science Chair and Distinguished Professor,
Indian Institute of Technology Hyderabad, Kandi, Talangana 502284, INDIA.
Email: m.vidyasagar@iith.ac.in.
This research is supported by the Science and Engineering Research Board (SERB),
India.
}
}

\date{\today}

\maketitle

\bc
\textbf{Dedicated to Prof.\ Eduardo Sontag on his 70th birthday, and \\
to Prof.\ Rajeeva L.\ Karandikar on his 65th birthday}
\ec

\begin{abstract}

In this paper, we study the almost sure boundedness and the convergence of the
stochastic approximation (SA) algorithm.
At present, most available convergence proofs are based on the
ODE method, and the almost sure boundedness of the iterations is an
assumption and not a conclusion.
In \cite{Borkar-Meyn00},
it is shown that if the ODE has only one globally attractive equilibrium,
then under additional assumptions, the iterations are bounded almost surely,
and the SA algorithm converges to the desired solution.
Our objective in the present paper is to provide an alternate proof
of the above, based on martingale methods,
which are simpler and less technical than those based on the ODE method.
As a prelude, we prove a new sufficient condition for the global asymptotic
stability of an ODE.
Next we prove a ``converse'' Lyapunov theorem
on the existence of a suitable Lyapunov function with a globally bounded
Hessian, for a globally exponentially stable system.
Both theorems are of independent interest to researchers in stability theory.
Then, using these results, we provide sufficient conditions for the 
almost sure boundedness and the convergence of the SA algorithm.
We show through examples that our theory covers some situations that are not
covered by currently known results, specifically \cite{Borkar-Meyn00}.

\end{abstract}

\section{Introduction}\label{sec:Intro}

The stochastic approximation (SA) algorithm, originally introduced 
by Robbins and Monro
\cite{Robbins-Monro51}, is a widely used method for finding a zero of
a function $\f : \R^d \ap \R^d$, when only noisy measurements of $\f$
are available.
Since its introduction, it has been a workhorse of probability theory
and has found many applications.
An early survey can be found in \cite{Lai03}, and a more recent
and broader survey can be found in \cite{Fran-Gram22}.

Here is a brief description of the algorithm.
Suppose $\f : \R^d \ap \R^d$, and it is desired
to find a solution $\bths$ to the equation $\f(\bth) = \bz$,
when one can access only noisy measurements of $\f(\cdot)$.
In broad terms, the algorithm proceeds as follows:
One begins with an initial guess $\bth_0$.
In the original version of SA proposed in \cite{Robbins-Monro51},
at time $t+1 \geq 1$, one has access to a noisy measurement
\be\label{eq:11a}
\y_{t+1} = \f(\bth_t) + \bxi_{t+1} ,
\ee
where $\bth_t$ is the current guess and $\bxi_{t+1}$ is the 
uncertain disturbance.
Then the guess $\bth_t$ is updated according to
\be\label{eq:11}
\bth_{t+1} = \bth_t + \al_t \y_{t+1} ,
= \bth_t + \al_t ( \f(\bth_t) + \bxi_{t+1} ),
\ee
where $\{ \al_t \}_{t \geq 1}$ is a prespecified deterministic
sequence of step sizes, with $\al_t \in (0,1)$ for all $t$.
It is desired to study the limit behavior of the sequence $\{ \bth_t \}$.
Define
\bd
S_f := \{ \bth \in \R^d : \f(\bth) = \bz \} .
\ed
The aims are to determine conditions under which (i) the sequence $\{ \bth_t \}$
is bounded almost surely, and (ii) 
$d(\bth_t,S_f) \ap 0$ as $\tai$,
where
\bd
d(\bth,S) := \inf_{\bphi \in S} \nmeu{\bth,\bphi} .
\ed

A well-established method for analyzing the limit behavior of $\{ \bth_t \}$
is known as the ODE method, whereby the sample paths of the recursion
\eqref{eq:11} are related to the (deterministic) solution trajectories of the
ODE $\dot{\bth} = \f(\bth)$.
This method was introduced in \cite{Der-Fradkov74,Ljung77b},
A very readable summary of this method is found in \cite{Meti-Priou84},
while book-length treatments of the method are available in
\cite{Kushner-Clark78,Benaim99,Borkar08}.
In the original paper \cite{Robbins-Monro51}, 
the step sizes $\al_t$ are positive, and satisfy the conditions below,
generally referred to as the Robbins-Monro conditions.
The two conditions are often shown together, but we display them separately
for reasons that will become clear later.
\be\label{eq:12a}
\sum_{t=1}^\infty \al_t^2 < \infty ,
\ee
\be\label{eq:12b}
\sum_{t=1}^\infty \al_t = \infty .
\ee

Until 2000, the available results had the following general format:
If the function $\f(\cdot)$ satisfies some  conditions, and
\textit{if} the sequence of iterates $\{ \bth_t \}$ is bounded almost surely,
\textit{then} $\bth_t \ap S_f$ almost surely, provided some other conditions
are satisfied.
Some authors refer to the almost-sure boundedness of the
iterates $\{ \bth_t \}$ as ``stability,'' and $\bth_t \ap S_f$ almost surely
as ``convergence.''
Thus the typical result stated that stability plus other conditions imply
convergence.
A major breakthrough was achieved in \cite{Borkar-Meyn00}, in which the
stability of the iterations is a conclusion and not a hypothesis.
Thus, under suitable conditions on the function $\f$, both stability and
convergence follow.
In that paper, it is assumed that there is a unique solution $\bths$
to $\f(\bth) = \bz$.
It is shown that, under suitable conditions (which are spelled out
precisely later; see Theorem \ref{thm:BM}), $\bth_t \ap \bths$
almost surely as $\tai$.

The present paper has the same objective, namely to make the stability
of the iterations a conclusion and not a hypothesis.
It is ironic that, as far back as 1965, there is a paper by Gladyshev
\cite{Gladyshev65}
that established both stability and convergence of the SA algorithm,
using martingale methods.
Moreover, there is a clear ``division of labor'' whereby \eqref{eq:12a}
leads to the almost sure boundedness of the iterations, while the
addition of \eqref{eq:12b} leads to convergence.
This division of labor is not found in any other paper until now.
However, the results in \cite{Gladyshev65}
are  restricted to a special class of functions $\f$.
Thus the present author was motivated by a desire to extend the
martingale methods of \cite{Gladyshev65}
to the same class of functions $\f$ as are covered by the ODE method
in \cite{Borkar-Meyn00}, or perhaps a more general class.
Subsequent to the publication of \cite{Gladyshev65},
Robbins and Siegmund \cite{Robb-Sieg71} introduced a
very general theorem that they called an ``almost supermartingale
convergence theorem,'' in which they rediscovered some of the basic ideas in
\cite{Gladyshev65}.
In \cite{Robb-Sieg71}, the authors mention that they were unaware
of \cite{Gladyshev65} at the time of writing their paper.

This is the starting point of the present paper.
Except in Theorem \ref{thm:20},
we study the case where $\bths$ is the unique solution of
$\f(\bth) = \bz$, and is thus the unique 
equilibrium of the associated ODE $\dot{\bth} = \f(\bth)$.
It is assumed that $\bths$ is a globally attractive equilibrium of the ODE.
The same assumptions are made in \cite{Borkar-Meyn00}.
Then we state and prove a new sufficient condition for global asymptotic
stability, namely Theorem \ref{thm:27},
which is less restrictive than the standard results
found in, for example, \cite{Hahn67,MV-93,Khalil02}.
The new sufficient condition allows us to conclude that the 
systems studied in \cite{Gladyshev65} are globally asymptotically stable --
a conclusion that does not follow from existing Lyapunov theory.
Then we present new sufficient conditions in Theorem \ref{thm:20}, for
the iterations $\{ \bth_t \}$ to (i) remain bounded almost surely, and
(ii) converge to $\bths$ almost surely.
These conditions require the existence of
a Lyapunov function that is bounded both above and below by multiples of
$\nmeusq{\bth}$, whose derivative can approach zero arbitrarily slowly.
The proof of Theorem \ref{thm:20}
is based on the Robbins-Siegmund theorem in \cite{Robb-Sieg71},
stated here as Theorem \ref{thm:RS}.
Then it is shown by example that Theorem \ref{thm:20} is applicable
to situations where the Borkar-Meyn result (Theorem \ref{thm:BM})
does not apply.
However, in Theorem \ref{thm:20}, the need to assume the
existence of a suitable Lyapunov function is still a bottleneck.
To overcome this, we strengthen the hypothesis to the requirement that
$\bths$ is a globally \textit{exponentially} stable equilibrium.
Then we state and prove a new converse theorem, namely Theorem \ref{thm:23},
on the existence of a
Lyapunov function that has a globally bounded Hessian matrix.
Combining these two theorems leads to a new ``self-contained'' result,
namely Theorem \ref{thm:21}, where all assumptions are only on $\f(\cdot)$.
The converse theorem proved here builds on an earlier theorem in
\cite{Corless-Glielmo98}, which uses a significantly different
Lyapunov function compared to its predecessors.
Both Theorems \ref{thm:27} and \ref{thm:23} are new and are possibly of
independent interest within the stability theory community.
In each case (namely, $\bths$ is either globally asymptotically stable,
or globally exponentially stable), the ``division of labor'' as in
\cite{Gladyshev65} continues to hold:
If \eqref{eq:12a} is satisfied,
then the sequence of iterates $\{ \bth_t \}$ is bounded almost surely.
Further, if \eqref{eq:12b} also holds, then $\bth_t \ap \bths$ almost surely.

The paper is organized as follows:
In Section \ref{sec:Main}, we introduce the list of various assumptions made,
and then state the main results of the paper.
For the convenience of the reader, we also state the relevant results
from \cite{Gladyshev65} and \cite{Borkar-Meyn00}.
This section also includes a discussion of the various assumptions and
the conclusions of the various theorems.
The actual proofs of the various theorems are given in Section \ref{sec:Proofs}.
Section \ref{sec:Exam} contains several illustrative examples that
highlight the advances made in our results when compared to known results.
Finally, Section \ref{sec:Conc} contains a discussion of some problems
for future research.

\section{Statements of Main Results}\label{sec:Main}

\subsection{List of Assumptions}

Throughout the remainder of the paper, 
within each theorem we choose from the following list of
``standing'' assumptions:
Note that not every assumption is needed in every theorem.

Let $\bth_0^t = (\bth_0 , \cdots , \bth_t)$,
$\bxi_1^t = (\bxi_1 , \cdots , \bxi_t)$, and let
$\F_t$ denote the $\s$-algebra generated by $\bth_0^t$, $\bxi_1^t$.
Then:
\ben
\item[(F1)] The equation $\f(\bth) = \bz$ has a unique solution $\bths$.
\item[(F2)] The function $\f$ is globally Lipschitz-continuous with
constant $L$.
\be\label{eq:21}
\nmeu{\f(\bth) - \f(\bphi)} \leq L \nmeu{\bth - \bphi} ,
\fa \bth , \bphi \in \R^d .
\ee
\item[(F2')] The function $\f$ is twice continuously differentiable, and
is globally Lipschitz-continuous with
constant $L$.
\item[(F3)] The equilibrium $\bths$ of the ODE $\dot{\bth} = \f(\bth)$ is
globally exponentially stable.
Thus there exist constants $\mu \geq 1, \g > 0$ such that
\be\label{eq:23}
\nmeu{\sb(t,\bth) - \bths} \leq \mu \nmeu{\bth - \bths} \exp(-\g t),
\fa t \geq 0 , \fa \bth \in \R^d .
\ee
\item[(F4)] There is a finite constant $K$ such that
\be\label{eq:24}
\nmS{\nabla^2 f_i(\bth)} \cdot \nmeu{\bth-\bths} \leq K ,
\fa i \in [d] , \fa \bth \in \R^d ,
\ee
where $[d]$ denotes the set $\{ 1 , \ldots , d \}$, and
$\nmS{\cdot}$ denotes the spectral norm of a matrix, i.e., its
largest singular value.
\item[(N1)] The noise sequence $\{ \bxi_t \}$ satisfies:
\be\label{eq:101}
E( \bxi_{t+1} | \F_t ) = \bz \as, \fa t \geq 0 .
\ee
\item[(N2)] The noise sequence $\{ \bxi_t \}$ satsifies:
\be\label{eq:102}
E( \nmeusq{\bxi_{t+1}} | \F_t ) \leq \s^2 (1 + \nmeusq{\bth_t - \bths} ) ,
\as \fa t \geq 0 ,
\ee
for some finite constant $\s^2$.
\een
Note that, as a consequence of Assumption (F2), for each $\bth \in \R^d$
there is a unique function $\sb(\cdot,\bth)$ that satisfies the ODE
\be\label{eq:22}
\dx{\sb(t,\bth)}{t} = \f(\sb(t,\bth)) , \sb(0,\bth) = \bth .
\ee
A consequence of Assumption (F4) is that
\be\label{eq:25}
\left| \tpx{ f_i(\bth)} { \th_j } {\th_k} \right |
\cdot \nmeu{\bth-\bths} \leq K ,
\fa i, j, k \in [d] , \fa \bth \in \R^d .
\ee

\subsection{Restatement of the Theorems of Gladyshev and Borkar-Meyn}

In this subsection, we restate the theorems of Gladyshev \cite{Gladyshev65}
and Borkar-Meyn \cite{Borkar-Meyn00} to facilitate comparison with 
the contributions of the present paper.

\begin{theorem}\label{thm:Glad}
(\cite{Gladyshev65}.)
Suppose assumptions (F1), (N1) and (N2) hold.
In addition, $\f(\cdot)$ is a passive function; that is, for each $0 < \e < M < \infty$,
\bd
\sup_{\e < \nmeu{\bth - \bths} < M} \IP{\bth - \bths}{\f(\bth)} < 0 ,
\ed
\bd
\nmeu{\f(\bth)} \leq K \nmeu{\bth-\bths} , K < \infty .
\ed
With these assumptions,
\ben
\item
If $\sum_{t=0}^\infty \al_t^2 < \infty$, then $\{ \bth_t \}$
is bounded almost surely.
\item If in addition $\sum_{t=0}^\infty \al_t = \infty$, then
$\bth_t \ap \bths$ almost surely as $\tai$.
\een
\end{theorem}

\textbf{Remark:}
Note that the second assumption implies that $\f(\cdot)$ is continuous
at $\bths$.
However, it need not be continuous anywhere else.

\begin{theorem}\label{thm:BM}
(\cite{Borkar-Meyn00}.)
Suppose assumptions (F1) and (N1) and (N2) hold, and in addition:
\bit
\item The Robbins-Monro conditions \eqref{eq:12a} and \eqref{eq:12b} hold.
\item $\f(\cdot)$ is globally Lipschitz continuous.
\item There is a ``limit function'' $\f_\infty : \R^d \ap \R^d$ such that
\bd
\frac{\f(r \bth)}{r} \ap \f_\infty(\bth) \mbox{ as } r \ap \infty ,
\ed
uniformly over compact subsets of $\R^d$.
\item $\bz$ is a globally exponentially stable
equilibrium\footnote{In \cite{Borkar-Meyn00}, only global asymptotic stability
is assumed.
However, as shown therein, because the vector field $\f_\infty$ is
``scale-free'' in that $\f_\infty(\l \bth) = \l \f_\infty (\bth)$
for all $\l > 0$, the two assumptions are equivalent.}
of
\bd
\dot{\bth} = \f_\infty(\bth) .
\ed
\eit
Under the stated assumptions,
\ben
\item $\{ \bth_t \}$ is bounded almost surely.
\item $\bth_t \ap \bths$ as $\tai$.
\een
\end{theorem}

At present, Theorem \ref{thm:BM} is \textit{the only} convergence result
based on the ODE method, that has the almost sure boundedness of the
iterations as a conclusion, and not a hypothesis.
The Lyapunov approach presented here is based on Assumption (F1),
namely that the equation $\f(\bth)$ has a unique solution $\bths$.
One of the perceived advantages of the ODE method is that it is
applicable even to the case where this equation has multiple solutions.
However, as shown in \cite{Meti-Priou84,Borkar08}, when the
equation under study has multiple solutions, the available results
once again have the almost sure boundedness of the iterations as
a hypothesis and not a conclusion.
The only result based on the ODE method where this is a conclusion and
not a hypothesis is \cite{Borkar-Meyn00}, and in this paper also it is
assumed that the equation $\f(\bth)$ has a unique solution.
In this sense, the results presented here are comparable to those
obtained using the ODE method.

\subsection{Preliminaries}\label{ssec:21}

The proofs of various theorems are
based on the following well-known theorem from \cite{Robb-Sieg71}.
The original reference \cite{Robb-Sieg71} is somewhat inaccessible.
However, the same theorem is stated as Lemma 2 in \cite[Section 5.2]
{BMP90}.
A recent survey of many results along similar lines is found in
\cite{Fran-Gram22}, where Theorem \ref{thm:RS} below is stated as Lemma 4.1.

\begin{theorem}\label{thm:RS}
(\cite{Robb-Sieg71}.)
Suppose $\{ z_t \} , \{ \eta_t \} , \{ \g_t \} , \{ \psi_t \}$ are
nonnegative stochastic processes adapted to some
filtration $\{ \F_t \}$, that satisfy
\be\label{eq:3311}
E(z_{t+1} | \F_t ) \leq (1 + \eta_t) z_t + \g_t - \psi_t \as, \fa t .
\ee
Define the set $\OM_0 \seq \OM$ by
\be\label{eq:3312}
\OM_0 := \{ \om : \sum_{t=0}^\infty \eta_t(\om) < \infty \}
\cap \{ \om : \sum_{t=0}^\infty \g_t(\om) < \infty \} ,
\ee
Then, for all\footnote{Here and elsewhere, ``for all'' really means
``for almost all.''} $\om \in \OM_0$, 
we have that (i) $\lim_{\tai} z_t(\om)$ exists and is finite, and (ii)
\be\label{eq:3313}
\sum_{t=0}^\infty \psi_t(\om) < \infty , \fa \om \in \OM_0 .
\ee
In particular, if $P(\OM_0) = 1$, then $\{ z_t \}$ is
bounded almost surely, in the sense that
\be\label{eq:3314}
P \{ \om \in \OM : \sup_t z_t(\om) < \infty \} = 1 ,
\ee
and
\be\label{eq:3315}
\sum_{t=0}^\infty \psi_t(\om) < \infty \as
\ee
\end{theorem}

Note that, while applying the above theorem here, it is always assumed that
$P(\OM_0) = 1$.

To state our first theorem on the convergence of the SA algorithm,
we begin by recalling some notation and concepts from nonlinear
stability theory.
The reader is referred to \cite{MV-93,Khalil02,Hahn67} for further details.

\begin{definition}\label{def:21}
A function $\phi : \R_+ \ap \R_+$ is said to \textbf{belong to class $\K$},
denoted by $\phi \in \K$, if $\phi(0) = 0$, and $\phi(\cdot)$ is 
strictly increasing.
A function $\phi \in \K$ is said to \textbf{belong to class $\KR$},
denoted by $\phi \in \KR$, if in addition, $\phi(r) \ap \infty$
as $r \ap \infty$.
A function $\phi : \R_+ \ap \R_+$ is said to \textbf{belong to class $\B$},
denoted by $\phi \in \B$, if $\phi(0) = 0$, and in addition, for all
$0 < \e < M < \infty$ we have that
\be\label{eq:109}
\inf_{\e \leq r \leq M} \phi(r) > 0 .
\ee
\end{definition}

See Example \ref{exam:0} for an example of a function that belongs to class
$\B$, but not to class $\K$.

Observe that if $\phi \in \KR$, then $\phi^{-1}$ exists and also belongs 
to $\KR$.

Next we recall a standard concept from stability theory.
Consider the ODE
\be\label{eq:110}
\dot{\bth} = f(\bth) ,
\ee
and suppose $V : \R^d \ap \R_+$ is $\C^1$.
Let $\gV : \R^d \ap \R^d$ denote the gradient of $V$, viewed as a
column vector.
Then the function $\Vd : \R^d \ap \R$ defined by
\be\label{eq:111}
\Vd(\bth) = \IP {\nabla V(\bth)}{ \f(\bth) }
\ee
is called the \textbf{derivative} of $V$ along the trajectories of
\eqref{eq:110}.

\subsection{Theorem Statements}\label{ssec:22}

We begin with
a theorem that gives a sufficient condition for global asymptotic stability.
The definition of global asymptotic stability is quite standard and can be
found in any standard reference, such as \cite{Hahn67,MV-93,Khalil02}.
Because Theorem \ref{thm:27} represents a weakening of the best-known sufficient
conditions, it should be of independent interest to researchers in nonlinear
stability theory.
Theorem \ref{thm:27} is the basis for extending
Theorem \ref{thm:Glad} to more general families of functions $\f(\cdot)$.
In the process, we are able to prove the convergence of the SA algorithm
in situations that are not covered by Theorem \ref{thm:BM}.

\begin{theorem}\label{thm:27}
Suppose Assumptions (F1) holds, and
that there exists a function $V: \R^d \ap \R_+$
and functions $\eta , \psi \in \KR, \phi \in \B$ such that
\be\label{eq:115}
\eta(\nmeu{\bth-\bths}) \leq V(\bth) \leq \psi(\nmeu{\bth-\bths})  ,
\fa \bth \in \R^d ,
\ee
\be\label{eq:116}
\Vd(\bth) \leq - \phi(\nmeu{\bth-\bths}) , \fa \bth \in \R^d ,
\ee
Then $\bths$ is a globally asymptotically stable equilibrium of the ODE
\eqref{eq:110}.
\end{theorem}

We now state the first result on
the almost-sure boundedness and convergence of the stochastic approximation
algorithm \eqref{eq:11}.
As is a recurring theme throughout the paper, the conditions for 
almost sure boundedness of the iterations are separate from 
and weaker than those for convergence.

\begin{theorem}\label{thm:20}
Suppose $\f(\bths) = \bz$, and Assumptions (F2) and (N1) and (N2) hold.
Suppose in addition that there exists a $\C^2$
Lyapunov function $V: \R^d \ap \R_+$
that satisfies the following conditions:
\bit
\item There exist constants $a , b > 0$ such that
\be\label{eq:112}
a \nmeusq{\bth-\bths} \leq V(\bth) \leq b \nmeusq{\bth-\bths} ,
\fa \bth \in \R^d .
\ee
\item There is a finite constant $M$ such that
\be\label{eq:114}
\nmS{\nabla^2 V(\bth)} \leq 2M , \fa \bth \in \R^d .
\ee
\eit
With these hypothesis, we can state the following conclusions:
\ben
\item If $\Vd(\bth) \leq 0$ for all $\bth \in \R^d$, and 
if \eqref{eq:12a} holds, then the iterations $\{ \bth_t \}$
are bounded almost surely.
\item Suppose further that there exists a function $\phi \in \B$ such that
\be\label{eq:113}
\Vd(\bth) \leq - \phi(\nmeu{\bth-\bths}) , \fa \bth \in \R^d ,
\ee
and in addition, both \eqref{eq:12a}, \eqref{eq:12b} hold.
Then $\bth_t \ap \bths$ almost surely as $\tai$.
\een
\end{theorem}

\begin{theorem}\label{thm:23}
Suppose Assumptions (F1), (F2'), (F3) and (F4) hold.
Under these hypotheses, there exists a $\C^2$ function $V : \R^d \ap \R_+$
such that $V$ and its derivative $\Vd : \R^d \ap \R$ defined by
\be\label{eq:28}
\Vd(\bth) = \IP {\nabla V(\bth)}{ \f(\bth) }
\ee
together satisfy the following conditions:
There exist positive constants $a, b, c$ and a finite constant
$M$ such that
\be\label{eq:29}
a \nmeusq{\bth-\bths} \leq V(\bth) \leq b \nmeusq{\bth-\bths} ,
\Vd(\bth) \leq -c \nmeusq{\bth - \bths} , \fa \bth \in \R^d .
\ee
\be\label{eq:210}
\nmS{\nabla^2 V(\bth)} \leq 2M , \fa \bth \in \R^d .
\ee
\end{theorem}

Combining Theorems \ref{thm:20} and \ref{thm:23} gives the following
``self-contained'' theorem:

\begin{theorem}\label{thm:21}
Suppose Assumptions (F1), (F2'), (F3) and (F4) as well as
Assumptions (N1)--(N2) hold.
Under these hypotheses,
\ben
\item If the step size sequence $\{ \al_t \}$ satisfies \eqref{eq:12a}, then
the stochastic process $\{ \bth_t \}$ is bounded almost surely.
\item If the step size sequence $\{ \al_t \}$ satisfies \eqref{eq:12b}
in addition to \eqref{eq:12a}, then
the stochastic process $\{ \bth_t \}$ converges almost surely to $\bths$
as $\tai$.
\een
\end{theorem}

\subsection{Discussion of the Theorems}\label{ssec:23}

\subsubsection{Theorem \ref{thm:27}}

The usual theorems on global asymptotic stability look similar to
Theorem \ref{thm:27},
except that the function $\phi(\cdot)$ in \eqref{eq:116} is assumed
to belong to the Class $\K$, not class $\B$; see for example
\cite[Theorem 26.2]{Hahn67} or \cite[Theorem 5.3.56]{MV-93}.
The change here is that the assumption on $\phi(\cdot)$ is weakened to
$\phi \in \B$ from $\phi \in \K$.
See Example \ref{exam:0} for a function that belongs to class $\B$ but
cannot be bounded below by any function of class $\K$.
Example \ref{exam:1} has a system whose global asymptotic stability
can be deduced using Theorem \ref{thm:27}, but not by traditional
theorems, using any Lyapunov function of the form $V(\th) = \th^{2m}, m \geq 1$.

\subsubsection{Theorem \ref{thm:20}}

\bit
\item
In the first part of the theorem that deals with the almost sure boundedness
of the iterations, there is \textit{no assumption} that $\bths$ is the unique
solution of $\f(\bth) = \bz$.
In other words, Assumption (F1) is \textit{not} made.
Moreover, the assumptions imply only that $\bths$ is a \textit{stable}
equilibrium of the ODE $\dot{\bth} = \f(\bth)$.
This is in sharp contrast to existing theorems in SA theory,
where it is assumed that $\bths$ is \textit{globally asymptotically stable}.
So far as the author is able to determine, there is no predecessor
to Theorem \ref{thm:20}, which establishes the almost-sure
boundedness of the iterations under a set of hypothesis that guarantee
\textit{only} that the equilibrium $\bths$ of the ODE $\dot{\bth} 
= \f(\bth)$ is stable -- not globally asymptotically stable.
\item
In the second part of the theorem that deals with the convergence of
the iterations, the assumption \eqref{eq:113} ensures that $\bths$
is a globally asymptotically stable equilibrium of the ODE $\dot{\bth}
= \f(\bth)$; this follows from Theorem \ref{thm:27}.
Therefore Assumption (F1) is implicit in the second part of the theorem.
\eit

\subsubsection{Theorem \ref{thm:23}}

Theorem \ref{thm:20} requires the existence of a suitable Lyapunov function $V$
that satisfies various conditions.
Therefore verifying whether or not such a function exists is a bottleneck.
It would be highly desirable to provide sufficient conditions that involve
only the function $\f(\cdot)$ that guarantee the existence of a suitable
Lyapunov function.
This is the objective of Theorem \ref{thm:23}.

As shown in Theorem \ref{thm:27}, the conditions
on $V$ in Theorem \ref{thm:20} ensure that the equilibrium
$\bths$ of the ODE \eqref{eq:110} is globally asymptotically stable.
By strengthening the assumption to the global \textit{exponential}
stability of $\bths$, and adding a few other assumptions, it is possible
to prove a ``converse'' Lyapunov theorem that establishes
the existence of a suitable $V$ function.
This is done in Theorem \ref{thm:23}.

Note that there is already a well-developed ``converse Lyapunov theory''
that establishes all the requirements on $V$, \textit{except} for the
global boundedness of the Hessian of $V$;
see for example \cite[Section 5.7]{MV-93} or \cite[Sections 48--51]{Hahn67}.
Therefore the contribution of Theorem \ref{thm:23} is in establishing
that $V$ has a globally bounded Hessian.
This theorem is new and possibly of independent interest to researchers in
nonlinear stability theory.

\subsubsection{Theorem \ref{thm:21}}

The closest available results to Theorem \ref{thm:21} in the current
literature are from \cite{Borkar-Meyn00}.
In that paper, it is also assumed that there is a unique solution $\bths$
to the equation $\f(\bth) = \bz$.
In addition, it is assumed that the functions
\bd
\f_r(\bth) : \bth \mapsto \frac{\f(r \bth)}{r} 
\ed
converge uniformly over compact sets to a limit function $\f_\infty$
as $r \ap \infty$,
and that $\bz$ is a globally exponentially stable equilibrium of the
ODE $\dot{\bth} = \f_\infty(\bth)$.
There is no requirement of the limit function $\f_\infty$ here.
Instead we have Assumption (F4), which
requires that the spectral norm of the Hessian matrix of each component
$f_i(\bth)$ decays at least as fast as $1/\nmeu{\bth-\bths}$.
It is now shown that Condition (F4) is only a slight strengthening
of the assumption that $\f(\cdot)$ is globally Lipschitz continuous.

\subsubsection{Significance of Assumption (F4)}

Suppose $\f: \R^d \ap \R^d$ is $\C^2$ and globally Lipschitz-continuous
with constant $L$.
Thus
\bd
\nmeu{ \f(\bth) - \f(\bphi) } \leq L \nmeu{ \bth - \bphi } ,
\fa \bth , \bphi \in \R^d .
\ed
Then $\f(\cdot)$ is absolutely continuous everywhere, and
it follows that the Jacobian $\nabla \f(\cdot) \in \R^{d \times d}$
exists almost everywhere is globally bounded, i.e.,
\bd
\nmS{\nabla \f(\bth)} \leq L , \fa \bth \in \R^d ,
\ed
where $\nmS{\cdot}$ is the spectral norm of a matrix, induced by
the $\ell_2$-vector norm.
Now observe that, for each index $i \in [d]$, we have
\begin{eqnarray*}
\nabla f_i(\bth) & = & \nabla f_i(\bths) 
+ \left[ \int_0^1 \nabla^2 f_i[ \bths + \l (\bth - \bths) ] \; d \l \right]
(\bth - \bths) \\
& = & \nabla f_i(\bths) + M_i(\bth,\bths) (\bth - \bths)  ,
\end{eqnarray*}
where
\bd
M_i(\bth,\bths) = \int_0^1 \nabla^2 f_i[ \bths + \l (\bth - \bths) ] \; d \l .
\ed
Hence it follows that
\be\label{eq:210a}
\nmS{ M_i(\bth,\bths) (\bth - \bths) } \leq 2L , \fa i \in [d],
\fa \bth , \bphi \in \R^d .
\ee
Now note that in effect Assumption (F4) consists of replacing the
integrand in the definition of $M_i(\bth,\bths)$ as follows:
\bd
\nmS{M_i(\bth,\bths)} \leftarrow \nmS{\nabla^2 f_i(\bth)} .
\ed
Almost all the literature on stochastic approximation assumes that
the function under study is globally Lipschitz-continuous.
In turn this imposes some restrictions on $\nabla^2 f(\cdot)$,
as shown in \eqref{eq:210a}.
The above argument shows Assumption (F4) is not too much stronger
than the consequence of the global Lipschitz continuity assumption.

\begin{example}\label{exam:5}
As a specific example, any vector field of the form
\bd
\f(\bth) = A (\bth - \bths) + \h(\bth) \nmeu{\bth - \bths}^{-2r}
\ed
would satisfy Assumption (F4) if (i) each component of $\h(\cdot)$ has a
globally bounded Hessian, and (ii) $r \geq 0.5$.
To see this, fix any index $i \in [d]$, and observe that
\bd
f_i(\bth) = \a^{(i)} (\bth - \bths) + h_i(\bth) \nmeu{\bth - \bths}^{-2r} ,
\ed
where $\a^{(i)}$ denotes the $i$-th row of $A$.
Thus
\bd
\px{f_i}{\th_j} = a_{ij} + \px{h_i}{\th_j}(\bth) \nmeu{\bth - \bths}^{-2r}
- 2r h_i(\bth) \nmeu{\bth - \bths}^{-2r-2} (\th_j - \th_j^*) ,
\ed
\begin{eqnarray*}
\tpx{f_i}{\th_j}{\th_k} & = & \tpx{h_i}{\th_j}{\th_k} \nmeu{\bth - \bths}^{-2r}
- 2r \px{h_i}{\th_j} (\th) \nmeu{\bth - \bths}^{-2r-2} (\th_k -\th_k^*)
- 2r \px{h_i}{\th_k} (\th) \nmeu{\bth - \bths}^{-2r-2} (\th_j - \th_j^*)\\
& + & 2r(2r+2) h_i(\bth) \nmeu{\bth - \bths}^{-2r-4} 
(\th_j - \th_j^*) (\th_k - \th_k^*)
- 2r h_i(\th) \nmeu{\bth - \bths}^{-2r-2} \d_{jk} ,
\end{eqnarray*}
where in the last equation $\d_{jk}$ denotes the Kronecker delta.
Now observe that, because $h(\cdot)$ has a globally bounded Hessian,
we have that
\bd
| [ \nabla^2 h(\bth) ]_{jk} | = O(1),
\nmeu{\nabla h_i(\bth)} = O(\nmeu{\bth-\bths}) ,
|h_i(\bth)| = O(\nmeusq{\bth-\bths}),
\ed
where $O(1)$ means bounded globally.
Thus every term in the expression for
$\partial^2 f_i/\partial \th_j \partial \th_k$
is $O(\nmeu{\bth - \bths}^{-2r})$.
Therefore
\bd
\nmS{\nabla^2 f_i(\bth)} = O(\nmeu{\bth - \bths}^{-2r}),
\nmS{\nabla^2 f_i(\bth)} \cdot \nmeu{\bth - \bths}
= O(\nmeu{\bth - \bths}^{-2r+1}) .
\ed
Since $-2r+1 \leq 0$ whenever $r \geq 0.5$, condition (F4) holds
whenever $r \geq 0.5$.
\halmos
\end{example}

The existence of a Lyapunov function $V$ that satisfies \eqref{eq:29}
is quite standard.
Indeed, the usual choice is
\be\label{eq:211}
V(\bth) := \int_0^\infty \nmeusq{\sb(t,\bth)} dt .
\ee
However, for this choice of $V$,
no conclusions can be drawn about the behavior of the gradient
$\nabla V$ nor the Hessian $\nabla^2 V$.
In \cite{Corless-Glielmo98}, the authors introduce a completely different
Lyapunov function of the form
\be\label{eq:212}
V(\bth) := \int_0^T e^{2\kappa \t} \nmeusq{\sb(\t,\bth) - \bths} \; d \t ,
\ee
where $0 < \kappa < \g$ is arbitrary, and $T$ is any \textit{finite}
number such that
\bd
\frac{\ln \mu}{\g - \kappa} \leq T < \infty ,
\ed
where $\mu, \g$ are defined in \eqref{eq:23}.
For this choice of Lyapunov function, it is shown in \cite{Corless-Glielmo98}
that there exists a finite constant $L'$ such that
\be\label{eq:213}
\nmeu{\nabla V(\bth)} \leq L' \nmeu{\bth-\bths} .
\ee
Now Theorem \ref{thm:23}
extends the theory further by showing that, if (F4) holds, then
$\nabla^2 V$ is globally bounded.
This in turn implies \eqref{eq:213}.
As we shall see, (F4) is the key assumption that allows us to extend the
converse Lyapunov theory of \cite{Corless-Glielmo98}
and prove that $V$ has a globally bounded Hessian.

\section{Proofs of the Theorems}\label{sec:Proofs}

\subsection{Proof of Theorem \ref{thm:27}}

\begin{proof}
Let $\bth(\cdot)$ denote a solution trajectory of the ODE \eqref{eq:110}.
Then \eqref{eq:116} implies that $V(\bth(t))$ is a nonincreasing
function of $t$, and therefore has a limit as $\tai$, call it $V_\infty$.
It is now shown that $V_\infty = 0$.
To see this, suppose that $V_\infty > 0$.
Then the right-side bound in \eqref{eq:115} implies that 
\bd
\nmeu{\bth-\bths} \geq \psi^{-1}(V_\infty) =: c_1 > 0 , \fa t ,
\ed
while the left-side bound in \eqref{eq:115} implies that
\bd
\nmeu{\bth-\bths} \leq \eta^{-1}(V(\bth(0))) =: c_2 < \infty .
\ed
In turn, this implies that 
\bd
- \Vd(\bth(t)) \geq \inf_{c_1 \leq r \leq c_2} \phi(r) > 0 ,
\ed
because $\phi(\cdot)$ is a function of class $\B$.
This is a contradiction because
\bd
V(\bth(t)) = V(\bth(0)) + \int_0^t \Vd(\bth(\t)) \; d\t ,
\ed
and if $- \Vd(\bth(\t))$ is bounded away from zero, then eventually
$V(\bth(t))$ would become negative.
Therefore $V_\infty = 0$, and $V(\bth(t)) \ap 0$ as $\tai$.
Now the left inequality in \eqref{eq:115} shows that $\nmeu{\bth(t)} \ap 0$
as $\tai$.
\end{proof}

\subsection{Proof of Theorem \ref{thm:20}\label{sec:3a}}

\begin{proof}
We begin by observing that, by Taylor's theorem, we have
\bd
V(\bth+ \boldeta) = V(\bth) + \IP{\gV(\bth)}{\boldeta}
+ \half \IP{\boldeta}{\nabla^2 V(\bth + \l \boldeta) \boldeta} ,
\ed
for some $\l \in [0,1]$.
Since $\nmS{\nabla^2 V(\bth + \l \boldeta)} \leq 2M$,
it follows that
\bd
V(\bth+ \boldeta) \leq V(\bth) + \IP{\gV(\bth)}{\boldeta}
+ M \nmeusq{\boldeta} , \fa 
\bth, \boldeta \in \R^d .
\ed
Now apply the above bound with $\bth = \bth_t$ and
$\boldeta = \al_t \f(\bth_t) + \al_t \bxi_{t+1}$.
This gives
\begin{eqnarray*}
V(\bth_{t+1}) & \leq & V(\bth_t) + \al_t \IP{\gV(\bth)}{\f(\bth_t)}
+ \al_t \IP{\gV(\bth_t)}{\bxi_{t+1} }
+ \al_t^2 M \nmeusq{ \f(\bth_t) + \bxi_{t+1} } \\
& = & V(\bth_t) + \al_t \Vd(\bth_t) + \al_t \IP{\gV(\bth_t)}{\bxi_{t+1} } 
+ \al_t^2 M [ \nmeusq{\f(\bth_t)} + \nmeusq{\bxi_{t+1}}
+ 2 \IP{ \f(\bth_t)} {\bxi_{t+1} } ] .
\end{eqnarray*}
Now we bound $E(V(\bth_{t+1}) | \F_t )$ using
Assumptions (F2), (N1), (N2) and the bounds in \eqref{eq:112}.
This gives
\be\label{eq:3315a}
E(V(\bth_{t+1}) | \F_t ) \leq V(\bth_t) 
+ \al_t^2 M [ L^2 \nmeusq{\bth_t - \bths} + \s^2 (1 + \nmeusq{\bth_t - \bths}) ] 
+ \al_t \Vd(\bth_t) .
\ee

To prove the first conclusion of the theorem, recall the hypotheses
that $\Vd(\bth) \leq 0$ for all $\bth$ and $\sum_{t=1}^\infty \al_t^2 < \infty$,
and apply \eqref{eq:112}.
Recall also that \eqref{eq:12a} holds.
Using these bounds in \eqref{eq:3315a} gives
\beq
E(V(\bth_{t+1}) | \F_t ) & \leq & V(\bth_t)
+ \al_t^2 M [ L^2 \nmeusq{\bth_t - \bths} + \s^2 (1 + \nmeusq{\bth_t - \bths}) ] 
\nonumber \\
& \leq & \left[ 1 + \al_t^2 \frac{M}{a} (L^2 + \s^2) \right] V(\bth_t)
+ \al_t^2 M \s^2 . \label{eq:3315b}
\eeq
Now apply Theorem \ref{thm:RS} with
\bd
z_t = V(\bth_t), \eta_t = \al_t^2 \frac{M}{a} (L^2 + \s^2) ,
\g_t = \al_t^2 M \s^2 , \psi_t = 0 .
\ed
Then it follows that $\lim_{\tai} V(\bth_t)$ exists almost surely and is finite.
Combined with \eqref{eq:112}, this shows that $\{ \bth_t \}$
is bounded almost surely.

To prove the second conclusion, we restore the term $\Vd(\bth_t)$
in \eqref{eq:3315a}, and use \eqref{eq:113}.
This gives
\bd
E(V(\bth_{t+1}) | \F_t  \leq \left[ 1 + \al_t^2 \frac{M}{a} (L^2 + \s^2)
\right] V(\bth_t) + \al_t^2 M \s^2  - \al_t \phi(\nmeu{\bth_t-\bths}) .
\ed
Now we again apply Theorem \ref{thm:RS} with
\bd
z_t = V(\bth_t), \eta_t = \al_t^2 \frac{M}{a} (L^2 + \s^2) ,
\g_t = \al_t^2 M \s^2 , \psi_t = \al_t \phi(\nmeu{\bth_t-\bths}) .
\ed
This time, the conclusions are that (i) there exists a random variable
$\zeta$ such that $V(\bth_t) \ap \zeta$ almost surely, and (ii)
\be\label{eq:120}
\sum_{t=0}^\infty \al_t \phi(\nmeu{\bth_t-\bths}) < \infty \as
\ee
Let $\OM_1 \seq \OM$ denote the values of $\om$ for which 
\bd
\sup_t V(\bth_t(\om)) < \infty , V(\bth_t(\om)) \ap \zeta(\om),
\mbox{ and } \sum_{t=0}^\infty \al_t \phi(\nmeu{\bth_t-\bths}) < \infty .
\ed
Note that $P(\OM_1) = 1$.
It is now shown that $\zeta(\om) = 0$ for all $\om \in \OM_1$.
Suppose by way of contradiction that, for some $\om \in \OM_1$,
we have that $\zeta(\om) = 2 \e > 0$.
Choose a $T$ such that $V(\bth_t(\om)) \geq \e$ for all $t \geq T$,
and also define $V_M := \sup_t V(\bth_t(\om))$.
Then we have that 
\bd
\sqrt{\e/b} \leq \nmeu{\bth_t} \leq \sqrt{V_M/a} , \fa t \geq T .
\ed
Define
\bd
\d := \inf_{\sqrt{\e/b} \leq r \leq \sqrt{V_M/a} } \phi(r) ,
\ed
and observe that $\d > 0$ because $\phi$ belongs to the class $\B$.
Therefore
\bd
\sum_{t=T}^\infty \al_t \phi(\nmeu{\bth_t-\bths})
\geq \sum_{t=T}^\infty \al_t \d = \infty ,
\ed
provided \eqref{eq:12b} holds.\footnote{Note that dropping
a finite number of terms does not affect the validity of \eqref{eq:12b}.}
But this contradicts \eqref{eq:120}.
Hence no such $\om \in \OM_1$ can exist.
In other words, $\zeta = 0$ almost surely, and $V(\bth_t) \ap 0$
almost surely.
Finally, it follows from \eqref{eq:112} that $\bth_t \ap \bths$
almost surely as $\tai$, which is the second conclusion.
\end{proof}

\subsection{Proof of Theorem \ref{thm:23}}\label{sec:3}

\begin{proof}
Following \cite{Corless-Glielmo98}, define the Lyapunov function candidate
$V$ as in \eqref{eq:212}.
Then, as shown in \cite{Corless-Glielmo98}, $V$ satisfies \eqref{eq:29}
and \eqref{eq:213}.
The latter is not of any concern to us.
So we focus on proving \eqref{eq:210}.

Note that the solution function $\sb(\cdot,\bth)$ satisfies
\be\label{eq:31}
\sb(t,\bth) = \bth + \int_0^t \f(\sb(\t,\bth)) \; d\t .
\ee
Therefore
\be\label{eq:31a}
\nabla_{\bth} \sb(t,\bth) = I + \int_0^t \nabla_{\bth} \f(\sb(\t,\bth)) \; d\t .
\ee
Next, the chain rule gives
\bd
\nabla_{\bth} \f(\sb(\t,\bth)) = \left. \nabla_{\bphi} \f(\bphi)
\right|_{\bphi = \sb(\t,\bth) } \nabla_{\bth} \sb(\t,\bth) .
\ed
Now the global Lipschitz continuity of $\f$ implies that
\bd
\nmS{ \nabla_{\bphi} \f(\sb(\t,\bphi)) } \leq L , \fa \bphi , \fa \t .
\ed
Therefore \eqref{eq:31a} leads to (after dropping the subscript $\bth$)
\bd
\nmS{ \nabla \sb(t,\bth) } \leq 1 + \int_0^t L \nmS{\nabla \sb(\t,\bth)}
\; d\t .
\ed
Now Gronwall's inequality leads to the bound
\be\label{eq:31b}
\nmS{ \nabla \sb(t,\bth) } \leq \exp(Lt) , \fa t , \fa \bth .
\ee

Next we proceed to find a bound on the second partial derivatives.
It follows from \eqref{eq:31} that
\bd
\px{s_i(t,\bth)}{\th_j} = \d_{ij} + \int_0^t \px{f_i(\sb(\t,\bth))}{\th_j}
\; d\t ,
\ed
where $\d_{ij}$ is the Kronecker delta.
Next,
\be\label{eq:32}
\tpx{s_i(t,\bth)}{\th_j}{\th_k} = 
\int_0^t \tpx{f_i(\sb(\t,\bth))}{\th_j}{\th_k} \; d\t .
\ee
We will use \eqref{eq:32} later.
Next, expand $V(\bth)$ as
\bd
V(\bth) = \int_0^T e^{2 \kappa \t} \sum_{i=1}^d [ s_i(\t,\bth) - \th_i^* ]^2 
\; d\t .
\ed
Thus
\bd
\px{V(\bth)}{\th_j} = \int_0^T 2 e^{2 \kappa \t} \sum_{i=1}^d 
[ s_i(\t,\bth) - \th_i^* ] \px{s_i(\t,\bth)}{\th_j} \; d\t ,
\ed
\bd
\tpx{V(\bth)}{\th_j}{\th_k}  = I_1 + I_2 ,
\ed
where
\be\label{eq:32a}
I_1 = \int_0^T 2 e^{2 \kappa \t} \sum_{i=1}^d
\px{s_i(\t,\bth)}{\th_k} \px{s_i(\t,\bth)}{\th_j} \; d\t ,
\ee
\be\label{eq:32b}
I_2 = \int_0^T 2 e^{2 \kappa \t} \sum_{i=1}^d
[ s_i(\t,\bth) - \th_i^* ] \tpx{s_i(\t,\bth)}{\th_j}{\th_k} \; d\t .
\ee
We will prove the boundedness of each integral separately.
Note that, as a consequence of \eqref{eq:31b}, we have
\bd
\left| \px{s_i(\t,\bth)}{\th_k} \right| ,
\left| \px{s_i(\t,\bth)}{\th_j} \right| \leq \nmS{\nabla \sb(\t,\bth) }
\leq \exp{L\t} , \fa \t , i, j, k .
\ed
So the first integral is bounded by
\bd
|I_1| \leq \int_0^T 2d e^{2 \kappa \t} e^{2L\t} \; d\t =: C_1 
< \infty 
\ed
for some constant $C_1$, whose precise value need not concern us.
So we concentrate on showing that, under Assumption (F4), $I_2$ is also
bounded globally.

Towards this end, we begin by observing that
\bd
\nmeu{\sb(t,\bth) - \bths} \geq e^{-Lt} \nmeu{\bth - \bths} , \fa t \geq 0 .
\ed
The proof is elementary and can be found in \cite[Theorem 8]{Brauer63}.
In particular
\be\label{eq:33}
\nmeu{\sb(t,\bth) - \bths} \geq e^{-LT} \nmeu{\bth - \bths} , \fa t \in [0,T] .
\ee
Now we estimate the entity $\partial^2 f_i(\sb(\t,\bth))/
\partial \th_j \partial \th_k$ in \eqref{eq:32}.
Note that
\bd
\px{f_i}{\th_j} ( \sb(\t,\bth)) = \sum_{l=1}^d
\left. \px{f_i(\bphi)}{\phi_l} \right|_{\bphi = \sb(\t,\bth)} 
\px{s_l(\t,\bth)}{\th_j} ,
\ed
\beq
\tpx{f_i(\sb(\t,\bth))}{\th_j}{\th_k} & = & \sum_{l=1}^d
\left. \px{f_i(\bphi)}{\phi_l} \right|_{\bphi = \sb(\t,\bth)}
\tpx{s_l(\t,\bth)}{\th_j}{\th_k} \nonumber \\
& + & \sum_{l=1}^d \px{}{\th_k} \left[ 
\left. \px{f_i(\bphi)}{\phi_l} \right|_{\bphi = \sb(\t,\bth)} \right]
\px{s_l(\t,\bth)}{\th_j} . \label{eq:34}
\eeq
The second term can be expanded as
\bd
\sum_{l=1}^d \left[ \sum_{r=1}^d
\left. \tpx{f_i(\bphi)}{\phi_l}{\phi_r} \right|_{\bphi = \sb(\t,\bth)}
\px{s_r(\t,\bth)}{\th_k} \right] \px{s_l(\t,\bth)}{\th_j}
\ed
Now Assumption (F4) and the bound \eqref{eq:33} together imply that
\bd
\left| \left. \tpx{f_i(\bphi)}{\phi_l}{\phi_r} \right|_{\bphi = \sb(\t,\bth)}
\right| \leq \nmS{\nabla^2 f_i(\sb(\t,\bth))} \leq
\frac{K}{\nmeu{\sb(\t,\bth) - \bths} }
\leq \frac{K e^{LT} }{\nmeu{\bth-\bths} } , \fa \t \in [0,T] .
\ed
Also, as shown in \eqref{eq:31b},
\bd
\left| \px{s_r(\t,\bth)}{\th_k} \right| ,
\left| \px{s_l(\t,\bth)}{\th_j} \right| \leq \nmS{\nabla \sb(\t,\bth)}
\leq e^{L\t} \leq e^{LT} , \fa \t \in [0,T].
\ed
Next, the global Lipschitz continuity of $\f$ implies that
\bd
\left| \px{f_i(\bphi)}{\phi_l} \right| \leq L .
\ed
Substituting all of these bounds including \eqref{eq:34}
into \eqref{eq:32} gives
\beq
\left| \tpx{s_i(t,\bth)}{\th_j}{\th_l} \right|
& \leq & \int_0^t L \sum_{l=1}^d \left| \tpx{s_l(\t,\bth)}{\th_j}{\th_k} \right|
\; d\t 
+ \int_0^t \sum_{l=1}^d \sum_{r=1}^d \frac{K e^{LT} e^{L\t} e^{L\t} }
{\nmeu{\bth-\bths} } \; d \t \nonumber \\
& \leq & C_2 + 
\int_0^t L \sum_{l=1}^d \left| \tpx{s_l(\t,\bth)}{\th_j}{\th_k} \right|
\; d\t , \label{eq:35}
\eeq
where
\bd
C_2 = \frac{d^2 TK e^{3LT}}{\nmeu{\bth-\bths} } 
\ed
is inversely proportional to $\nmeu{\bth-\bths}$.
Now define
\bd
h_{jk}(t,\bth) := \sum_{i=1}^d \left| \tpx{s_i(t,\bth)}{\th_j}{\th_k} \right| .
\ed
Note that the right side of \eqref{eq:35} does not depend on $i$.
Therefore \eqref{eq:35} implies that
\begin{eqnarray*}
h_{jk}(t,\bth) & \leq & \sum_{i=1}^d \left[ C_2 + \int_0^t L \left|
\sum_{l=1}^d \tpx{s_i(\t,\bth)}{\th_j}{\th_k} \right| \right] \\
& \leq & C_2 d + \int_0^t Ld h_{jk}(\t,\bth) \; d\t .
\end{eqnarray*}
So by Gronwall's inequality
\bd
h_{jk}(t,\bth) \leq C_2 d e^{LdT} , \fa t \in [0,T] .
\ed
Since $h_{jk}$ is a sum, each individual component must also be smaller than
$h_{jk}$ in magnitude.
Thus
\bd
\left| \tpx{s_i(t,\bth)}{\th_j}{\th_l} \right| \leq C_2 d e^{LdT} 
\leq \frac{C_3}{ \nmeu{\bth-\bths} }
\ed
for a suitable constant $C_3$.
Therefore we have established that the Hessian of each $s_i$ decays
as $\bth$ gets farther away from $\bths$.
Now we return to $I_2$ as defined in \eqref{eq:32b}, and observe that,
as a consequence of Assumption (F3) of global exponential stability, we have
\bd
| s_i(t,\bth) - \th_i^* | \leq \nmeu{\sb(t,\bth) - \bths}
\leq \mu \nmeu{\bth-\bths} , \fa t \geq 0 .
\ed
Now in the definition of $I_2$, we get the bound
\bd
| s_i(t,\bth) - \th_i^* | \cdot 
\left| \tpx{s_i(t,\bth)}{\th_j}{\th_k} \right| 
\leq \mu \nmeu{\bth-\bths} \cdot \frac{C_3}{ \nmeu{\bth-\bths} }
= \mu C_3 .
\ed
Since the integrand in \eqref{eq:32b} is bounded and $T$ is finite, it
follows that $I_2$ is also bounded.
This finally leads to the desired conclusion that $\nmS{ \nabla^2 V}$
is globally bounded.
\end{proof}

Note that in the above proof, the finiteness of the constant $T$ is crucial.
Therefore the traditional infinite integral type of Lyapunov function
defined in \eqref{eq:211} is not directly amenable to such analysis.
It is perhaps possible to replace the Lyapunov function candidate
of \eqref{eq:211} by another function of the form
\bd
V(\bth) := \left[ \int_0^\infty \nmm{\sb(t,\bth)}^{2p} \; dt \right]^{1/p} .
\ed
However, something similar to Assumption (F4) would still be required.

\section{Examples}\label{sec:Exam}

In this section we give a few examples of the results presented thus far.
We also discuss the implications of Assumption (F4) in \eqref{eq:24}.

\begin{example}\label{exam:0}
Observe that every $\phi$ of class $\K$ also belongs to class $\B$.
However, the converse is not true.
Define
\bd
\phi(r) = \left\{ \ba{ll} r, & \mbox{if } r \in [0,1] , \\
e^{-(r-1)}, & \mbox{if } r > 1 . \ea \right.
\ed
Then $\phi$ belongs to Class $\B$.
However, since $\phi(r) \ap 0$ as $r \ap \infty$, $\phi$ cannot be
bounded below by any function of class $\K$.
\end{example}

\begin{example}\label{exam:1}
This example illustrates how Theorem \ref{thm:27} goes beyond
currently available theorems in Lyapunov stability theory.
The current theorems, of which \cite[Theorem 26.2]{Hahn67} and
\cite[Theorem 5.3.56]{MV-93} are typical,
require $- \Vd(\th)$ to be a function of Class $\K$.

Recall the function $\phi(\cdot)$ defined in Example \ref{exam:0}.
Now consider the ODE $\dot{\th} = f(\th)$, where
\bd
f(\th) = \left\{ \ba{ll} - \phi(\th) , & \th \geq 0 , \\
-f(-\th) , & \th < 0 . \ea \right.
\ed
Thus $f(\cdot)$ is just an odd extension of $-\phi(\cdot)$.
If we choose the Lyapunov function $V(\th) = \th^2$, then
\bd
\Vd(\th) = - | \th | \cdot \phi(|\th|) .
\ed
Therefore $-\Vd$ is a function of class $\B$, and the global asymptotic
stability of the equilibrium $\th^* = 0$ follows from Theorem \ref{thm:27}.
However, $- \Vd(\th)$ is \textit{not} a function of class $\K$, nor can
it be bounded below by a function of Class $\K$ because
$|\th|\cdot\phi(|\th|) \ap 0$ as $|\th| \ap \infty$.
More generally,
for \textit{every} function of the form $V(\th) = \th^{2m} , m \geq 1$,
$-\Vd$ cannot be bounded below by a function of Class $\K$.
Hence the traditional theorems fail to apply for any such function $V$.
\end{example}

\begin{example}\label{exam:2}
Using Theorem \ref{thm:Glad}, one can infer the convergence of
the SA algorithm in the one-dimensional case, when the measurement
$y_{t+1}$ is of the form \eqref{eq:11a}, with $f(0) = 0$, and
\be\label{eq:61}
\sup_{\e \leq \th \leq M} \th f(\th) < 0 , \fa 0 < \e < M < \infty .
\ee
Such a function $f(\cdot)$ is called a ``passive'' function in circuit theory.
The objective of this example is to demonstrate that Gladyshev's
result does \textit{not} in general follow from those in 
\cite{Borkar-Meyn00}, but \textit{does} follow from Theorem \ref{thm:20}.

Consider the one-dimensional ODE $\dot{\th} = - f(\th)$,
where $f(0) = 0$ and satisfies \eqref{eq:61},
$|f(\th)| \ap 0$ as $|\th| \ap \infty$.
Then $f(\cdot)$ satisfies the hypotheses of Theorem \ref{thm:Glad},
In this case, the scale-free function defined in \cite{Borkar-Meyn00},
namely
\bd
f_\infty ( \th ) := \lim_{r \ap \infty} \frac{f(r \th)}{r} \equiv 0 , \fa \th .
\ed
Hence the ODE $\dot{\th} = f_\infty(\th)$ cannot be globally
asymptotically state.
On the other hand, if we use the Lyapunov function $V(\th) = \th^2$
(which is in effect what is done in \cite{Gladyshev65}), then
\bd
- \Vd(\th) = - \th f(\th) ,
\ed
which is a function of Class $\B$.
Hence it follows from Theorem \ref{thm:20} that $\{ \th_t \}$ is
bounded almost surely if \eqref{eq:12a} holds.
If in addition, \eqref{eq:12b} also holds, then $\th_t \ap 0$
almost surely as $\tai$.

Note that, whenever $\f(\cdot)$ remains bounded as $\nmeu{\bth} \ap 0$,
we get $\f_\infty \equiv \bz$ for all $\bth$.
Hence Theorem \ref{thm:BM} cannot be applied to such a situation.
\end{example}

\begin{example}\label{exam:2a}
As an illustration of Theorem \ref{thm:20}  for the case where
$\f(\bth) = \bz$ has multiple solutions,
consider the following function $f: \R \ap \R$:
\bd
f(\th) = \left\{ \ba{ll} - 1 + \sin (\th + \pi/2) , & \th \geq 0 , \\
- f( \th) , & \th < 0 .
\ea \right.
\ed
The solutions of this equation are $\th_n = 2 \pi n$, for every integer $n$.

Now define $\th^* = 0$ to be the solution of interest to us,
and define the Lyapunov function $V(\th) = \th^2$.
Then $\Vd(\th) = \th f(\th) \leq 0$ for all $\th$.
Therefore all assumptions of \textit{the first part} of Theorem
\ref{thm:20} are satisfied, and we can infer that the iterations
$\{ \th_t \}$ are almost surely bounded whenever the step sizes
$\{ \al_t \}$ are square summable.

It is worth noting that the equilibrium $\th^* = 0$ is asymptotically
stable but not globally asymptotically stable.
\end{example}

\begin{example}\label{exam:3}
A standard problem in Reinforcement Learning is known as ``value
evaluation,'' wherein one wants to solve a linear equation of the form 
\bd
\v^* = \rbold + \g A \v^* ,
\ed
where $\v^* \in \R^d$ is called the ``value'' vector,
$\rbold \in \R^d$ is called the ``reward'' vector, $\g \in (0,1)$ is
called the ``discount factor,'' and $A \in \R^{d \times d}$
is the state transition matrix of a Markov chain.
Hence, if we define
\bd
\nmm{M}_{\infty \ap \infty} := \sup_{\v \neq \bz} 
\frac{\nmi{M \v}}{\nmi{\v}} ,
\ed
then  $\nmm{A}_{\infty \ap \infty} = 1$, and
$\nmm{\g A}_{\infty \ap \infty} = \g < 1$.

To apply SA to this problem, let us switch notation to be consistent
with that in the paper, and rewrite as
\bd
\bth = \rbold + \g A \bth .
\ed
If we define the function $\f$ via
\bd
\f(\bth) = \rbold + \g A \bth - \bth ,
\ed
then the unique equilibrium of the associated ODE $\dot{\bth} = \f(\bth)$
is indeed the desired solution $\v^*$.
If it happens that the $\ell_2$-induced norm $\nmS{\g A} > 1$,
then there exists a $\v \in \R^d$ such that
\bd
\IP{ \v - \v^* }{ \f(\v) - \f(\v^*) } > 0 .
\ed
Thus the map $\f$ does not satisfy the hypotheses of Theorem \ref{thm:Glad}.
However, the convergence of the SA algorithm can still be inferred
using Theorem \ref{thm:20}, as follows:
Note that, since $\g < 1$ and $\r(A) \leq 1$, the eigenvalues of the
matrix $\g A - I$ all have negative parts.
Hence $\v^*$ is a globally exponentially stable equilibrium.
Then it follows from \cite[Theorem 5.4.42]{MV-93} that, whenever
$Q$ is a symmetric, positive definite matrix, the so-called Lyapunov
matrix equation
\bd
P ( \g A - I) + (\g A - I)^\top P = - Q
\ed
has a unique positive definite solution for $P$.
Thus $V(\bth) = \bth^\top P \bth$ satisfes the hypotheses of
Theorem \ref{thm:20} (because the Hessian of $V(\cdot)$ is constant
and thus bounded).
Hence we can conclude that the SA algorithm will converge to the desired
solution $\v^*$, provided \eqref{eq:12a} and \eqref{eq:12b} hold.
\end{example}

\section{Conclusions and Future Work}\label{sec:Conc}

In this paper, we have presented some simple proofs for the 
almost sure boundedness and convergence of
the stochastic approximation (SA) algorithm, based on martingale methods and
converse Lyapunov theory.
Two new results have been presented in Lyapunov stability:
The first is a new sufficient condition for global asymptotic stability,
which is weaker than currently known conditions.
The second is a ``converse'' theorem that ensures the existence of
a Lyapunov function with a globally bounded Hessian matrix for
globally exponentially stable systems.
Each of these theorems is coupled with the well-known
Robbins-Siegmund theorem to provide
some simple proofs for the convergence of
the stochastic approximation (SA) algorithm.
The results presented here in Lyapunov theory are new and may be of
independent interest to researchers in nonlinear stability theory.
The fact that the convergence proofs of SA are
based on Lyapunov theory, and not the ODE
method discussed in \cite{Der-Fradkov74,Ljung77b,Kushner-Clark78},
opens the possibility that the same approach can be used to prove
the convergence of the SA algorithm in more general settings, such as
two-time scale SA \cite{CL-SB-Auto17} or projected gradient SA
\cite{Tsi-Van-TAC97}.
These lines of research are currently under investigation.

A different class of stochastic algorithms is studied
in \cite{Ljung78}.
Specifically, the basic recursion is
\be\label{eq:51}
\bth_{t+1} = \bth_t + \al_t ( \f(\bth_t) + \bxi_{t+1} + \boldeta_{t+1} ) ,
\ee
where the function $\f(\cdot)$ is a \textit{gradient vector field}, that is,
\bd
\f(\bth) = - \nabla J(\bth)
\ed
for some function $J : \R^d \ap \R$ with compact level sets.
This means  that, for every constant $c \in \R$, the level set
\bd
S_J(c) := \{ \bth \in \R^d : J(\bth) \leq c \}
\ed
is compact.
In addition, there are now \textit{two} measurement noise sequences
$\{ \bxi_t \}$ and $\{ \boldeta_t \}$.
The sequence $\{ \boldeta_t \}$ converges to zero almost surely as $\tai$,
while the sequence $\{ \bxi_t \}$ \textit{is not required to satisfy} the
conditional zero mean assumption \eqref{eq:101}.
There are other minor differences, but these are the main differences
between the set-up studied here and that studied in \cite{Ljung78}.
A similar extension is also presented in \cite[p.\ 17]{Borkar08}.
It is worth noting that, in this more general setting,
the almost sure boundedness
of the iterations is \textit{assumed and not inferred}.
See for example Assumption (A4) on \cite[p.\ 11]{Borkar08}.
Our current research includes an extension of the martingale method
to the case where Assumptions (N1) and (N2) on the noise (namely 
\eqref{eq:101} and \eqref{eq:102}) are \textit{not} assumed.
In particular, the noise is not assumed to have zero conditional mean,
and the constant $\s^2$ is replaced by a time-varying number $\s^2_t$
that is allowed to be unbounded.
Despite these relaxations, the almost sure boundedness of the iterations
is inferred and not assumed, and the convergence is established.
Those results will be presented elsewhere.

In \cite{BMP90}, specifically Chapter 1 of Part II, the authors
introduce a class of SA algorithms that are more general than those
studied here.
In broad terms, the iterations are driven by a Markov process.
Specifically (see \cite[Eq.\ (1.1.1)]{BMP90}), the general formulation,
converted to the present notation, is
\be\label{eq:52}
\bth_{t+1} = \bth_t + \al_t H(\bth_t,X_{t+1}) + \al_t^2 \r_t (\bth_t,X_{t+1}) ,
\ee
where $\bth_t \in \R^d$ and $X_t \in \R^n$.
Assumption (A2) in \cite{BMP90} then requires that there exists a family
$\pi_\bth(x,A)$ of transition probabilities 
on $\R^n$ such that, for every Borel subset $A \seq \R^n$, we have
\bd
\Pr \{ X_{t+1} \in A | \F_t \} = \pi_{\bth_t} ( X_t,A) ,
\ed
where $\F_t$ is the $\s$-algebra generated by $\bth_0^t, X_1^t$.
The authors observe that if $\pi_\bth(x,dz) = \mu_\bth(dz)$ for
some probability measure $\mu$ (that is, the transition probability
does not depend on $x$), then the above formulation reduces to the
Robbins-Monro formulation studied here.
It would be of interest to explore whether martingale-based methods
can be extented to this more general situation.

As a final comment, it might be possible to apply Lyapunov methods to
establish the almost sure boundedness of the iterations, and then to
use ODE methods to derive more detailed estimates about the convergence
than is possible using Lyapunov methods.
This approach merits further study.

\section*{Acknowledgements}

The author thanks Prof.\ Rajeeva Karandikar for assistance in understanding
the contents of \cite{Gladyshev65},
Prof.\ Boris Polyak for drawing his
attention to the reference \cite{Fran-Gram22}, and to  Prof.\ Barbara
Franci for providing a copy of the difficult to locate paper
\cite{Robb-Sieg71}.
The author also thanks one of the reviewers for an extremely thorough
and helpful review that substantially improved the paper.

\section*{Conflict of Interest}

The author does not declare any conflict of interest.


\end{document}